\documentclass[lettersize,journal]{IEEEtran}
\usepackage{amsmath,amsfonts}
\usepackage{algorithmic}
\usepackage{algorithm}
\usepackage{array}
\usepackage[caption=false,font=normalsize,labelfont=sf,textfont=sf]{subfig}
\usepackage{textcomp}
\usepackage{stfloats}
\usepackage{multirow}
\usepackage{url}
\usepackage{verbatim}
\usepackage{graphicx}
\usepackage{cite}
\usepackage{threeparttable}
\usepackage{color} 

\begin{document}

\title{A Spatial-Channel-Temporal-Fused Attention for Spiking Neural Networks}

\author{Wuque Cai,
	    Hongze Sun,
	    Rui Liu,
	    Yan Cui,
	    Jun Wang,
	    Yang Xia,
	    Dezhong Yao, \emph{Senior Member, IEEE}, 
	    and Daqing Guo
        % <-this % stops a space
\thanks{This work was supported in part by the STI 2030--Major Project under Grant 2022ZD0208500 and in part by the National Natural Science Foundation of China under Grant 61933003, Grant 82072011, and Grant 31771149. (Corresponding authors: Dezhong Yao; Daqing Guo).

Wuque Cai, Hongze Sun, Rui Liu, Jun Wang, Yang Xia, Daqing Guo are with the Clinical Hospital of Chengdu Brain Science Institute, MOE Key Lab for NeuroInformation, School of Life Science and Technology, University of Electronic Science and Technology of China, Chengdu 611731, China (e-mail: dqguo@uestc.edu.cn).

Yan Cui is with the Department of Neurosurgery, Sichuan Provincial People's Hospital, University of Electronic Science and Technology of China, Chengdu 610072, China.

Dezhong Yao is with 
the Clinical Hospital of Chengdu Brain Science Institute, MOE Key Lab for NeuroInformation, School of Life Science and Technology, University of Electronic Science and Technology of China, Chengdu 611731, China, also with the Research Unit of NeuroInformation (2019RU035), Chinese Academy of Medical Sciences, Chengdu 611731, China, and also with the School of Electrical Engineering, Zhengzhou University, Zhengzhou 450001, China (e-mail: dyao@uestc.edu.cn).}
% <-this % stops a space
}

\maketitle

\begin{abstract}
	Spiking neural networks~(SNNs) mimic brain computational strategies, and exhibit substantial capabilities in spatiotemporal information processing. As an essential factor for human perception, visual attention refers to the dynamic process for selecting salient regions in biological vision systems. Although visual attention mechanisms have achieved great success in computer vision applications, they are rarely introduced into SNNs. Inspired by experimental observations on predictive attentional remapping, we propose a new spatial-channel-temporal-fused attention (SCTFA) module that can guide SNNs to efficiently capture underlying target regions by utilizing accumulated historical spatial-channel information in the present study. Through a systematic evaluation on three event stream datasets (DVS Gesture, SL-Animals-DVS and MNIST-DVS), we demonstrate that the SNN with the SCTFA module (SCTFA-SNN) not only significantly outperforms the baseline SNN (BL-SNN) and two other SNN models with degenerated attention modules, but also achieves competitive accuracy with existing state-of-the-art methods. Additionally, our detailed analysis shows that the proposed SCTFA-SNN model has strong robustness to noise and outstanding stability when faced with incomplete data, while maintaining acceptable complexity and efficiency. Overall, these findings indicate that incorporating appropriate cognitive mechanisms of the brain may provide a promising approach to elevate the capabilities of SNNs.
\end{abstract}


\section{Introduction}
\IEEEPARstart{O}{ver} the past decade, spiking neural networks (SNNs) have received unprecedented attention and have exhibited powerful capabilities in various intelligent scenarios~\cite{maass1997networks, ghosh2009spiking, tavanaei2019deep}, such as pattern recognition~\cite{yu2015spiking, cheng2020lisnn, zhang2021event, hu2018spiking}, object tracking~\cite{pei2019towards, kim2020spiking} and robot navigation~\cite{rast2018behavioral, bing2018survey}. In SNNs, neural information is represented and transmitted in the form of discrete spike events. However, such non-differentiable feature of spikes makes the learning in SNNs to be more difficult than that in conventional artificial neural networks (ANNs). To overcome this challenge, several training approaches, such as gradient-based optimization~\cite{wu2018spatio, wu2019direct} and ANN-to-SNN conversion~\cite{rueckauer2017conversion, rathi2021diet, wu2021tandem}, have been proposed to develop SNNs with competitive performance. A deeper understanding of the computational principles of realistic brain can inspire us to design more efficient SNN models.

Visual attention is a critical cognitive process that allows humans to selectively allocate limited attention resources to underlying targets~\cite{carrasco2011visual, lamme2003visual}. This biological mechanism is essential for visual processing, because it enables individuals to respond quickly and accurately to relevant stimuli in their environment while filtering out irrelevant distractions. Remarkably, recent years have witnessed a great development of visual attention in computer vision, and various types of attention modules in channel~\cite{hu2018squeeze, WangWZLZH20}, spatial~\cite{jaderberg2015spatial, wang2018non, shen2019visual}, temporal~\cite{xu2017jointly, zhang2019scan} and their combined dimensions~\cite{woo2018cbam, song2017end, fu2020scene} have been designed for different neural network models. Although visual attention has achieved great success in traditional ANN models, attention-related mechanisms have rarely been considered in SNNs. Despite this fact, we noted that a few recent studies demonstrated the potential of visual attention in improving the performance of SNNs~\cite{yao2021temporal, zhu2022tcja, 103389fnins20221079357, kundu2021spike, liu2022event, liu2022general}. However, to the best of our knowledge, most of these studies concentrated on attention mechanisms in one or two dimensions, and it still remains challenging to efficiently combine and fuse spatial, channel and temporal information altogether with a flexible approach.

Recent experimental studies have highlighted the existence of a predictive attentional remapping mechanism in the mammalian brain~\cite{melcher2007predictive, mathot2010evidence, rolfs2011predictive, szinte2018pre, wilmott2021transsaccadic, golomb2021visual}. By measuring the spatiotemporal change in attention when saccades occur, it has been widely observed that attention can be shifted to the underlying target area before the eyes move toward it~\cite{melcher2007predictive, rolfs2011predictive, mathot2010evidence, golomb2021visual}. Further experiments on attention remapping have demonstrated that accurate predictive remapping highly relies on the spatial and temporal dynamics of attention and requires sufficient accumulation of spatiotemporal information~\cite{szinte2018pre, wilmott2021transsaccadic}. From a functional perspective, these findings provide an efficient approach for predictively shifting attention to potential areas based on historical information. Inspired by these experimental findings, we suggest that incorporating the mechanism of predictive attentional remapping into SNNs with attention modules may be helpful for locating underlying target regions. To mimic  predictive attentional remapping in the brain, we first capture the spatial-channel information of spiking features by developing a high-dimensional attention tensor. We then accumulate historical information by embedding this information into the intrinsic temporal dynamics of neurons in SNNs. Theoretically, our bioinspired design concept is feasible because an appropriate accumulation of historical spatial-channel information can improve the empirical feature representation capability of SNNs.

Accordingly, in this study, we propose a novel spatial-channel-temporal-fused attention (SCTFA) module for SNNs. For this purpose, we employ a three-dimensional (3-D) spatial-channel attention block to capture discriminative spatial-channel information of underlying targets. Then, we introduce the feedback of attention into the temporal evolution of the membrane potentials of neurons (see Methods for details). With this method, the attention tensor has a long-term effect on neuronal membrane potentials, and its historical spatial-channel information is regulated by the decay factor of membrane potentials. For simplicity, we name the SNN with the SCTFA module as SCTFA-SNN. From a functional perspective, the major novelty of our SCTFA-SNN model is that it can make full use of the accumulated spatial-channel information to determine underlying target regions. To examine the performance of the SCTFA-SNN model, we evaluated and compared it with other SNN models on three benchmark event-stream datasets. Our detailed experiments and analyses clearly demonstrate the superiority of the proposed SCTFA-SNN model in event stream classifications, with our model achieving competitive accuracy with existing state-of-the-art (SOTA) methods.

The main contributions of our work are summarized as follows:

\begin{itemize}
	\item[$\bullet$] We design an SCTFA module to extract and accumulate spatial-channel information in the temporal (historical) domain. Accordingly, the proposed SCTFA module can make full use of spatial, channel and temporal information to locate targets.
	\item[$\bullet$]  We introduce a long-term effect of the attention tensor on neuronal membrane potentials. With this approach, we can realize the concept of predictive remapping of attention with the help of historical spatial and channel information.
	\item[$\bullet$] Ablation study shows that the SCTFA-SNN outperforms the baseline SNN (BL-SNN) and two other SNN models with degenerated spatial-temporal-fused and channel-temporal-fused attention modules.	
	\item[$\bullet$]  The SCTFA-SNN model achieves better or comparable accuracy with other SOTA models on three benchmark event stream datasets and exhibits substantial superiority in terms of robustness and stability to noisy and incomplete data.
\end{itemize}

The remainder of this paper is structured as follows. In the next section, we review recent developments in training methods and attention mechanisms for SNNs, and introduce  typical strategies for improving the representation of event streams. Then, we describe the neuron model, the structure of the SCTFA module and the learning method for training the SCTFA-SNN models in Section III. Section IV provides the detailed experimental results and analysis of the SCTFA-SNN model on different datasets. Finally, we present a brief conclusion in Section V.

\section{Related Work}
\subsection{Training Methods for SNNs}
There are various strategies for SNN training, and ANN-to-SNN conversion and gradient-based optimization are two fundamental approaches~\cite{rueckauer2017conversion, rathi2021diet, wu2021tandem, wu2018spatio, wu2019direct, deng2021comprehensive}. As an indirect supervised learning approach, ANN-to-SNN conversion involves first training an ANN and then mapping the parameters to an SNN with an equivalent architecture~\cite{rueckauer2017conversion, rathi2021diet, wu2021tandem}. For a sufficient inference time, the performance of converted SNNs can achieve near-lossless accuracy. On the other hand, gradient-based optimization is a promising supervised learning approach that directly trains SNNs by calculating the gradient through backpropagation (BP). However, because the spike activity is non-differentiable, the true gradient cannot be accurately computed; therefore, surrogate and approximate gradients are commonly introduced to address this problem~\cite{wu2018spatio, wu2019direct}. To date, several spike-based BP methods have been developed. Typical methods include, but are not limited to, the spike layer error reassignment in time (SLAYER)~\cite{shrestha2018slayer}, the spatiotemporal BP (STBP)~\cite{wu2018spatio, deng2021comprehensive}, and the BP through time (BPTT)~\cite{werbos1990backpropagation}. Among them, the STBP approach is believed to be a powerful learning method for highly efficient SNN models, and this method is also employed in our study to train different SNN models.

\subsection{Attention Mechanisms in SNNs}
In computer vision, many types of attention models and modules have been developed to capture underlying regions of interest while neglecting other irrelevant information~\cite{hu2018squeeze, WangWZLZH20, jaderberg2015spatial, shen2019visual, wang2018non, xu2017jointly, zhang2019scan, woo2018cbam, song2017end, fu2020scene}. However, compared with ANNs, only a few studies have attempted to incorporate attention mechanisms in SNN models~\cite{yao2021temporal, zhu2022tcja, 103389fnins20221079357, kundu2021spike, liu2022event, liu2022general}. For example, Yao et al. designed a temporal-wise attention (TA) module for input frames to obtain their statistical characteristics at different timesteps~\cite{yao2021temporal}. With this method, the importance of different frames can be judged during training and the irrelevant frames can be discarded at the inference stage, thus resulting in substantially improved performance. On the basis of the TA module, a temporal-channel joint attention (TCJA) architectural unit was developed for establishing the association between the timestep and the channel~\cite{zhu2022tcja}. The resulting TCJA-SNN model has been demonstrated to exhibit significant superiority on several mainstream static and neuromorphic datasets. Inspired by the dynamical features of biological synapses, a recent study proposed a spatio-temporal synaptic connection SNN (STSC-SNN) model consisting of two plug-and-play blocks, a temporal response filter (TRF) and a feedforward lateral inhibition (FLI), to enhance the temporal information processing capabilities of SNNs~\cite{103389fnins20221079357}. Additionally, several other studies further demonstrated the effectiveness of visual attention by introducing different types of spatial-wise, channel-wise and temporal-wise attentions into SNN models~\cite{kundu2021spike, liu2022event, liu2022general}. Motivated by  predictive attentional remapping~\cite{rolfs2011predictive, mathot2010evidence, szinte2018pre, wilmott2021transsaccadic, golomb2021visual}, we construct a spatial-channel-temporal-fused attention module for SNNs in this study. The proposed SCTFA module allows SNNs to effectively locate underlying target regions by using  accumulated historical spatial-channel information. Unlike the TA and TCJA modules, our SCTFA module does not choose important input frames, but makes full use of historically-accumulated spatial-channel information in the temporal direction. Accordingly, the SCTFA module provides a flexible approach to deeply integrate spatial, channel, and temporal information for SNNs.

\subsection{Representation of event streams}
We use several event stream datasets recorded by event cameras to assess our models~\cite{amir2017low, vasudevan2022sl, serrano2015poker}. As a biological-inspired sensor, an event camera responds to light intensity changes in each sensor pixel with high temporal resolution, low energy consumption and high dynamic range~\cite{serrano2013128, vasudevan2020introduction}. For event streams, information is encoded with sparsely asynchronous spatiotemporal events. To improve the information representation of event streams, one mainstream approach is to split event-based data into different slices and convert these data into the frame-based form. Although frame-based representation discards the sparsity of asynchronous events, this approach considerably reduces the difficulty of the learning process. Due to the relatively high signal-to-noise ratio, the frame-based representation has been widely used to process event streams~\cite{lee2016training, wu2022brain, fang2021incorporating, wu2021liaf, yao2021temporal, zheng2021going, he2020comparing, kugele2020efficient}. Furthermore, several data augmentation strategies have been recently proposed as a new approach to enhance the representation ability of sparse events. Typical examples are EventMix~\cite{shen2022eventmix} and EventDrop~\cite{gu2021eventdrop}. However, for a fair comparison with other SOTA results based on SNN models, we employ the frame-based representation for event streams and do not consider any data augmentation strategies in this work.

%%%%%%%%%%%%%%%%%%%%%%%%%%%%%%Fig1
\begin{figure*}[!t]
	\centering
	\includegraphics{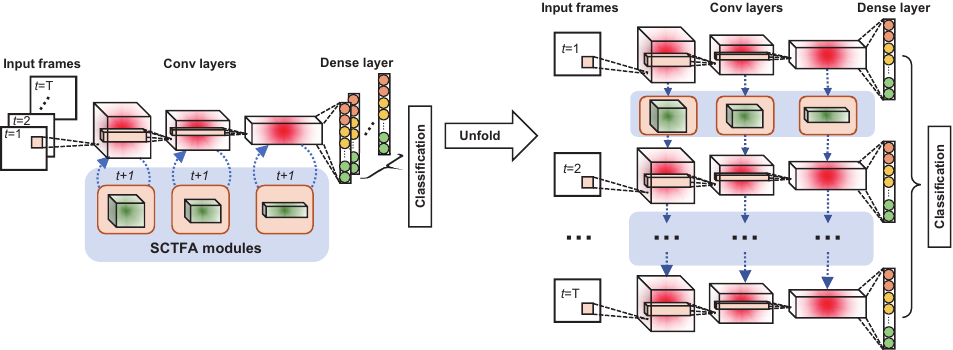}
	\caption{The architecture and unfold form of the proposed SCTFA-SNN model. The SCTFA module is inserted into each convolutional (Conv) layer and converts the spiking feature maps (SFMs) into an attention tensor (blue). The SFMs extracted by the last convolutional layer are input into a dense layer for classification.}
	\label{fig1}
\end{figure*}
%%%%%%%%%%%%%%%%%%%%%%%%%%%%%%

\section{Methods}
In this section, we first describe the leaky-integrate-and-fire~(LIF) neuron model used in this study, and then introduce the structure of the SCTFA module and the learning method of the SCTFA-SNN model in detail.
	
\subsection{LIF Neuron Model}
A large variety of models have been developed to resemble the spiking dynamics of biological neurons. The LIF neuron model, which is one of the most commonly used models, allows us to simulate the dynamics of SNNs in an efficient way~\cite{gerstner2014neuronal}. The differential equation of a single LIF model can be described as follows:
\begin{equation}
	\tau\frac{\text{d} v_i}{\text{d} t} = -\left(v_i - v_{\text{rest}}\right) + I_i,
	\label{Eq1}
\end{equation}
where $v_i$ is the membrane potential of the $i$-th neuron, $\tau$ is the membrane time constant, $v_{\text{rest}}$ is the resting potential, and $I_i$ is the input current from its presynaptic neurons. Whenever the membrane potential of a neuron reaches a threshold at $v_{\text{th}}$, a spike is generated and after that its membrane potential is reset to the resting potential $v_{\text{rest}}$.  Mathematically, this process produces a binary spike train $s_i=(s_i^{1}, ..., s_i^{t}, ..., s_i^{T})$ for the $i$-th neuron, with:	
\begin{equation}
	\label{Eq2}
	s_i^t = g(v_i) = 
	\begin{cases}
		1  & \mbox{if }v_i\ge v_{\text{th}},\\
		0  & \text{otherwise}.
	\end{cases}
\end{equation}

Without loss of generality, we set $v_{\text{rest}}=0$~mV and derive a simple iterative form of the LIF neuron model as follows:
\begin{equation}
	v_i^{t+1} = \kappa_{\tau} v_i^{t}\left(1-s_i^{t}\right) + \frac{\text{d} t}{\tau}I_{i}(t),
	\label{Eq3} 
\end{equation}
where $v_i^{t}$ and $I_{i}(t)$ are the membrane potential and synaptic input of the $i$-th neuron at time $t$, $\text{d} t$ represents the integration timestep, and $\kappa_{\tau}=1-\frac{\text{d} t}{\tau}$ is a decay factor. By eliminating the scaling effect $\frac{\text{d} t}{\tau}$ into the synaptic weights, we obtain the final iterative formula of the LIF neuron model as follows:
\begin{equation}
	v_i^{t+1} = \kappa_{\tau} v_i^{t}\left(1-s_i^{t}\right) + \sum_{j}w_{ij}s_j^{t+1},
	\label{Eq4} 
\end{equation}
where the outer sum runs over all synapses onto the $i$-th neuron, and $w_{ij}$ is the scaled synaptic weight from the $j$-th neuron to the $i$-th neuron.

\subsection{Structure of the SCTFA Module}
We employ SNNs with convolutional structures and the frame-based representation for event stream processing. As seen in the network architecture [Fig.~\ref{fig1}~(left)], we insert an SCTFA module into each convolutional layer to convert the spiking feature maps (SFMs). For each convolutional layer, the SFMs at time $t$ are converted into an attention tensor that affects the same layer at time $t+1$. This attention module transfers spatial-channel information through temporal dimension, which constitutes the basic structure of the SCTFA module. The SFMs extracted by the last convolutional layer are input into a dense layer for voting with a rate-based decoding strategy. For a clearer understanding, the unfolded structure of the SCTFA-SNN model is also schematically depicted in Fig.~\ref{fig1}~(right).

%%%%%%%%%%%%%%%%%%%%%%%%%Fig 2
\begin{figure*}[t]
	\centering
	\includegraphics{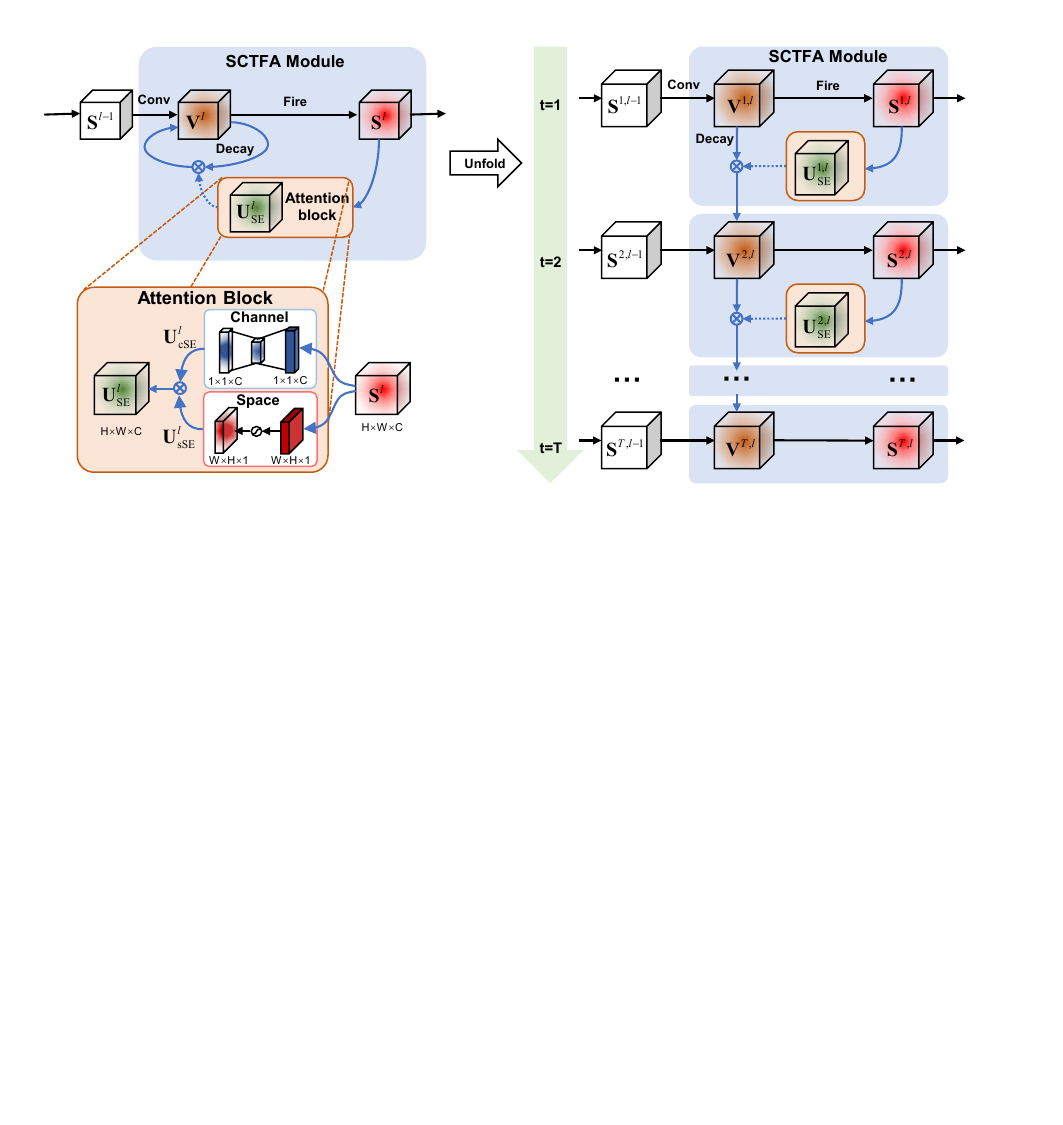}
	\caption{Diagram of the SCTFA module. The attention block in the SCTFA module is composed of both channel and spatial attention blocks. This attention block generates a 3-D attention tensor that excites corresponding neurons in the same layer, with its historical influence adjusted by the decay factor of the membrane potentials.}
	\label{fig2}
\end{figure*}
%%%%%%%%%%%%%%%%%%%%%%%%%

In the SCTFA module, the SFMs are transferred by an attention block to generate an attention tensor with the same size [Fig.~\ref{fig2}]. To do this, a spatial-channel attention block (3-D), consisting of a spatial attention block (2-D) and a channel attention block (1-D), is designed to extract information from the output SFMs~\cite{roy2018concurrent, woo2018cbam}. Mathematically, we define the output SFMs of the $l$-th convolutional layer at time $t$ represented  by the binary spiking tensor $\mathbf{S}^{t,l}$ with size of ${H \times W \times C}$, where $C$ is the number of channels, and $H$ and $W$ are the height and width of the SFMs. In this study, the spatial attention block is generated by channel squeeze and spatial excitation through the convolution and nonlinear transformation of the SFMs as follows~\cite{roy2018concurrent}:
\begin{equation}
	\mathbf{U}_{\text{sSE}}^{t,l} = \sigma\left(\mathbf{W}_{\text{s}}^{l}\ast \mathbf{S}^{t,l} + b^{l}\right),
	\label{Eq5}
\end{equation}
where $\sigma(\cdot)$ is the sigmoid function, $\mathbf{W}_{\text{s}}^{l} \in \mathbb{R}^{1 \times 1 \times C \times 1}$ and $b^{l} \in \mathbb{R}$ are the weight matrix of the kernel and the bias vector of the convolution, respectively, and $\mathbf{U}_{\text{sSE}}^{t,l}\in \mathbb{R}^{1 \times H \times W}$ is the spatial attention tensor representing the relative weight in the spatial direction. On the other hand, we use the channel attention mechanism to explore the inter-channel relationship of SFMs. Unlike the spatial attention block that learns `where' to focus on, the channel attention is established under the assumption that different feature maps have different importance levels and the channel attention block is designed to efficiently capture important feature maps. In this study, the channel attention block is acquired by spatial squeeze and channel excitation. For this purpose, we first perform a squeeze operation in the spatial direction:
\begin{equation}
	\mathbf{e}^{t,l} = \frac{1}{H \times W}\sum_{i=1}^{H}\sum_{i=1}^{W}\mathbf{S}^{t,l}(i, j),
	\label{Eq6}
\end{equation}
where $\mathbf{e}^{t,l} \in \mathbb{R}^{C \times 1 \times 1}$ is the average channel vector in the spatial direction. Then, channel excitation can be achieved by using two fully connected nonlinear transformations as follows~\cite{hu2018squeeze}: 
\begin{equation}
	\mathbf{U}_{\text{cSE}}^{t,l} = \sigma\left( \mathbf{W}_{\text{c2}}^{l} \cdot  \delta\left(\mathbf{W}_{\text{c1}}^{l}\mathbf{e}^{t,l}\right) \right),
	\label{Eq7}
\end{equation}
where $\delta(\cdot)$ is the ReLU function, $\mathbf{W}_{\text{c1}}^{l} \in \mathbb{R}^{\frac{C}{r} \times C}$ and $\mathbf{W}_{\text{c2}}^{l} \in \mathbb{R}^{C \times \frac{C}{r}}$ are the weights of the linear function with a reduction ratio of $r$, and  $\mathbf{U}_{\text{cSE}}^{t,l} \in \mathbb{R}^{C \times 1 \times 1}$ is the channel attention tensor denoting the relative weight in the channel direction. As previously discussed in~\cite{hu2018squeeze}, the reduction ratio $r$ is a hyperparameter that is used to control the trade-off between performance and computational cost of the channel attention block in the model. In principle, a smaller $r$ may endow the model with better performance but increases the number of parameters. However, a too small $r$ might lead to overfitting, which may also reduce the model performance. Based on these considerations, we choose an intermediate value of $r=4$ in the present work to ensure that the model has a good balance between accuracy and complexity.

We generate an overall 3-D attention tensor by combining above spatial and channel attention tensors as follows:
\begin{equation}
	\mathbf{U}_{\text{SE}}^{t,l}=\mathbf{U}_{\text{sSE}}^{t,l}\left(\mathbf{S}^{t,l}\right) \odot \mathbf{U}_{\text{cSE}}^{t,l}\left(\mathbf{S}^{t,l}\right),
	\label{Eq8}
\end{equation}
where $\odot$ is the Hadamard product, and $\mathbf{U}_{\text{SE}}^{t,l} \in \mathbb{R}^{C \times H \times W}$ denotes the 3-D attention tensor of the SFMs. Each element within this tensor corresponds to a neuron in the SFMs, and its value represents the importance level of the neuron at time $t$. To achieve a long-term influence process, we integrate the feedback of the 3-D attention tensor into the membrane potentials of neurons and rewrite the iterative formula of the LIF neuron model in the SCTFA-SNN model as follows:
\begin{equation}
	v_i^{t+1,l}=\kappa_{\tau} v_i^{t,l} u^{t,l}_{\text{SE};i} \left(1 - s_i^{t,l}\right) + \sum_{j}w_{ij}^{l,l-1}s_j^{t+1,l-1},
	\label{Eq9}
\end{equation}
where $s^{t,l}_{i}$ and $v^{t,l}_{i}$ are the binary spike and membrane potential of the $i$-th neuron in the $l$-th layer at time $t$,  $u^{t,l}_{\text{SE};i} \in \mathbf{U}_{\text{SE}}^{t,l}$ is the corresponding attention value, and $w_{ij}^{l,l-1}$ is the synaptic weight from the $j$-th neuron in the $(l-1)$-th layer to the $i$-th neuron in the $l$-th layer. Theoretically, the decay factor of the neuronal membrane determines the impact of historical information on the current influence. From a functional perspective, the SCTFA-SNN model uses the accumulated historical spatial-channel information to capture the underlying target regions.

\subsection{Learning Method}
In this study, we use a rate-based decoding strategy for event stream classification. To this end, we consider the last layer in the SNN as the voting layer, which is composed of $M$ classes with each class represented by $P$ neurons. For a given input with the $m$-th class, we encourage neurons representing this class to have the highest spiking activity. Thus, the output vector $\mathbf{o}^{t} = \{o_1^t, o_2^t,\cdots,o_m^t,\cdots,o_M^t\}$ at time $t$ can be calculated as follows:
\begin{equation}
	o_{m}^{t}=\frac{1}{P}\sum_{i=(m-1)P}^{mP}s_{i}^{t,L},
	\label{Eq10}
\end{equation}
where $o_m^t$ is the average spiking events of neurons in the $m$-th class at time $t$. Then, the loss function is described by the mean squared error as follows:
\begin{equation}
	Loss = \frac{1}{2N}\sum_{n=1}^{N}\left\| \mathbf{Y}_{n}-\frac{1}{T}\sum_{t=1}^{T}\mathbf{o}_{n}^{t} \right\|_{2}^{2}.
	\label{Eq11}
\end{equation}
Here $\mathbf{o}_{n}^{t}$ represents the average voting vector at timestep $t$, $\mathbf{Y}_{n}$ denotes the label vector of input data, $N$ is the number of training samples, and $T$ is the total number of timesteps in the simulation.

We use the STBP method to train the SNN models ~\cite{wu2019direct, wu2018spatio}. In the STBP method, the gradient of the loss function $L$ with respect to $s^{t,l}_{i}$ and $v^{t,l}_{i}$ can be calculated as follows:
\begin{equation}
	\frac{\partial{Loss}}{\partial{s^{t,l}_{i}}}=\sum_{j=1}^{n^{l+1}}\frac{\partial{Loss}}{\partial{s^{t,l+1}_{j}}}\frac{\partial{s^{t,l+1}_{j}}}{\partial{s^{t,l}_{i}}}+\frac{\partial{Loss}}{\partial{s^{t+1,l}_{i}}}\frac{\partial{s^{t+1,l}_{i}}}{\partial{s^{t,l}_{i}}},
	\label{Eq12}
\end{equation}
\begin{equation}
	\frac{\partial{Loss}}{\partial{v^{t,l}_{i}}}=\frac{\partial{Loss}}{\partial{s^{t,l}_{i}}}\frac{\partial{s^{t,l}_{i}}}{\partial{v^{t,l}_{i}}}+\frac{\partial{Loss}}{\partial{s^{t+1,l}_{i}}}\frac{\partial{s^{t+1,l}_{i}}}{\partial{v^{t,l}_{i}}}.
	\label{Eq13}
\end{equation}
In this study, a surrogate function is utilized to approximate the activation of discrete spikes ~\cite{wu2019direct, wu2018spatio}:
\begin{equation}
	g(v^{t,l}_{i}) = \frac{1}{\pi}\text{arctan}\left[\frac{\pi}{2}\alpha (v^{t,l}_{i}-v_{\text{th}})\right] + \frac{1}{2}, 
	\label{Eq14}
\end{equation}
where a factor of $\alpha=2$ is used to adjust the slope of the surrogate function near zero. 

However, learning in the SCTFA module cannot be directly accomplished with the standard STBP method~\cite{wu2019direct, wu2018spatio}. When inputting the SFMs of the $l$-th layer at time $t$ into the SCTFA module, an attention tensor is introduced to adjust the membrane potentials of neurons in the $l$-th layer at time $t+1$. In this case, the gradient in the SCTFA module with respect to $u_{\text{SE};i}^{t,l}$ and $s^{t,l}_{i}$ can be written as follows:
\begin{equation}
	\frac{\partial{s^{t+1,l}_{i}}}{\partial{u^{t,l}_{\text{SE};i}}}=\frac{\partial{s^{t+1,l}_{i}}}{\partial{v^{t+1,l}_{i}}}\kappa_{\tau}v^{t,l}_{i}\left(1-s^{t,l}_{i}\right),
	\label{Eq15}
\end{equation}
\begin{equation}
	\frac{\partial{u^{t,l}_{\text{SE};i}}}{\partial{s^{t,l}_{i}}} =u^{t,l}_{\text{sSE};i}\frac{\partial{u^{t,l}_{\text{cSE};i}}}{\partial{s^{t,l}_{i}}}+u^{t,l}_{\text{cSE},i}\frac{\partial{u^{t,l}_{\text{sSE};i}}}{\partial{s^{t,l}_{i}}}.
	\label{Eq16}
\end{equation}
With Eqs.(\ref{Eq11})-(\ref{Eq16}), we can efficiently train the proposed SCTFA-SNN model [see Algorithm~\ref{Alg1}].  By removing the channel or spatial block from the SCTFA module, our model degenerates into an SNN with spatial-temporal-fused attention (STFA-SNN) or channel-temporal-fused attention (CTFA-SNN).

%%%%%%%%%%%%%%%%%%%%%%%%algorithm
\begin{algorithm}[t]
	\small 
	\caption{Training of the SCTFA-SNN model.}\label{Alg1}
	\begin{algorithmic}
		\REQUIRE Input sample $\left\{\mathbf{S}^{t,0} \right\}_{t=1}^{T}$, target label $Y$, parameters of synaptic weights and SCTFA modules in each layer~$(\mathbf{W}^{l}, \mathbf{W}_{c1}^{l},\mathbf{W}_{c1}^{l},\mathbf{W}_{s}^{l}, \mathbf{b}^{l})_{l=1}^{L}$, states of the membrane potential and spike train of each layer~$(\mathbf{V}^{t,l}, \mathbf{S}^{t,l})$, spike counter of SNN $\mathbf{o}^{L}$, and 1-D average pooling (AP) //~Calculating network output from voting layer. 
		\ENSURE Updated parameter $\mathbf{W}^{l}, \mathbf{W}_{c1}^{l},\mathbf{W}_{c1}^{l},\mathbf{W}_{s}^{l}~\text{and}~\mathbf{b}^{l}$
		\FOR{$l=1$ to $L+1$}
		\STATE $\mathbf{V}^{1,l}$,$\mathbf{S}^{1,l} \gets \text{StateUpate}\left(\mathbf{V}^{1,l},\mathbf{S}^{1,l-1},\mathbf{W}^{l}\right)$// Eqs.~(\ref{Eq2})-(\ref{Eq4})
		\ENDFOR
		\FOR{$t=2$ to $T$}
		\FOR{$l=1$ to $L+1$}
		\IF{ConvLayer}
		\STATE $\mathbf{U}_{\text{SE}}^{t,l} \gets \text{FusedAttention}\left(\mathbf{U}_{\text{sSE}}^{t,l},\mathbf{U}_{\text{sSE}}^{t,l}\right)$ // Eqs.~(\ref{Eq5})-(\ref{Eq8})
		\STATE $(\mathbf{V}^{t,l}$,$\mathbf{S}^{t,l}) \gets \text{StateUpate}\left(\mathbf{S}^{t,l-1},\mathbf{V}^{t,l},\mathbf{S}^{t-1,l},\mathbf{U}_{\text{SE}}^{t,l},\mathbf{W}^{l}\right)$  // Eq.~(\ref{Eq9})
		\ELSE
		\STATE $(\mathbf{V}^{t,l}$,$\mathbf{S}^{t,l}) \gets \text{StateUpate}\left(\mathbf{V}^{t,l},\mathbf{S}^{t-1,l},\mathbf{W}^{l}\right)$
		\ENDIF
		\ENDFOR
		\STATE $\mathbf{o}^{L} \gets \mathbf{o}^{L} + \mathbf{S}^{t,L}$ //~Calculating the number of voting spikes.
		\ENDFOR
		\STATE $\mathbf{p} \gets \text{AP}\left(T,\mathbf{o}^{L}\right)$  // Eq.~(\ref{Eq10})
		\STATE $Loss \gets \text{MSE}\left (\mathbf{Y}, \mathbf{p}\right)$ // Eq.~(\ref{Eq11})
		\STATE \text{Update} $\{\mathbf{W}^{l}, \mathbf{W}_{c1}^{l},\mathbf{W}_{c1}^{l},\mathbf{W}_{s}^{l},\mathbf{b}^{l}\}_{l=1}^{L}$ //~Eqs.~(\ref{Eq11})-(\ref{Eq16})
	\end{algorithmic}
	\label{alg1}
\end{algorithm}
%%%%%%%%%%%%%%%%%%%%%%%

\section{Experiments}
\subsection{Datasets and Implementations}

In this study, we test our models on three benchmark event stream datasets, including two gesture recognition datasets (DVS Gesture~\cite{amir2017low} and SL-Animals-DVS~\cite{vasudevan2022sl}) and one converted DVS dataset (MNIST-DVS~\cite{serrano2015poker}). All these datasets were recorded by event cameras, and have a spatial resolution of 128$\times$128 pixels, with the two channels representing positive and negative changes in the illumination of the scene. For each event stream data, we split it into different slices with fixed length $\Delta t$, and choose the first $T$ slices as training and test data. Then, we accumulate spike trains for each slice and feed them into the SNNs. The detailed descriptions of these event stream datasets are given as follows.

%%%%%%%%%%%%%%%%%%%%%%%%%%% Table1
\begin{table}[t]
	\renewcommand{\arraystretch}{1.2}
	\newcommand{\tabincell}[2]{
		\begin{tabular}
			{@{}#1@{}}#2
		\end{tabular}
	}

	\caption{Default values of hyperparameters for different datasets.}
	\label{Tab1}
	\centering
	\setlength{\tabcolsep}{2.5mm}{
		\begin{tabular}{|c|c|c|c|}
			\hline
			      Dataset        & DVS Gesture & SL-Animals-DVS & MNIST-DVS \\ \hline
			       Epoch         &     200     &      200       &    100    \\
			     Batch Size      &     16      &       25       &    100     \\
			       $\eta$        &    0.002    &     0.0002     &   0.001   \\
			      $\gamma$       &    0.98     &      0.97      &   0.95    \\
			$v_{\text{th}}$~(mV) &    1.15     &      1.5       &    0.8    \\
			  $\kappa_{\tau}$    &     0.7     &      0.4       &    0.5    \\
			        $r$          &      4      &       4        &     4     \\
			        $T$          &     10      &       30       &    20     \\
			  $\Delta t$~(ms)    &     125     &       50       &    25     \\
			      $L$~(ms)       &    1250     &      1500      &    500    \\ \hline
		\end{tabular}}
\end{table}
%%%%%%%%%%%%%%%%%%%%%%%%%%%

%%%%%%%%%%%%%%%%%%%%%%%%%Fig 3
\begin{figure*}[t]
	\centering
	\includegraphics{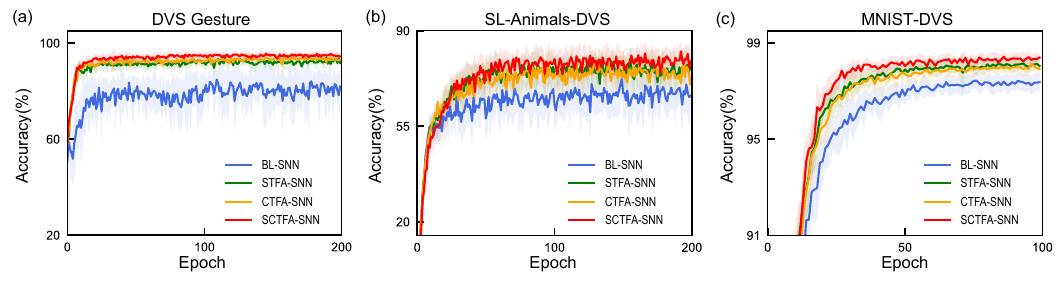}
	\caption{Average training curves of different SNN models on three datasets. (a) DVS Gesture, (b) SL-Animals-DVS and (c) MNIST-DVS. Different colors represent different SNN models: BL-SNN (blue), STFA-SNN (green), CTFA-SNN (orange) and SCTFA-SNN (red). }
	\label{fig3}
\end{figure*}
%%%%%%%%%%%%%%%%%%%%%%%%%

%%%%%%%%%%%%%%%%%%%%%%%%%%% Table2
\begin{table}[t]
	\renewcommand{\arraystretch}{1.2}
	
	\newcommand{\tabincell}[2]{
		\begin{tabular}
			{@{}#1@{}}#2
		\end{tabular}
	}
	
	\caption{Architectures of SNN models for different datasets.}

	\centering
	\setlength{\tabcolsep}{2mm}{
		\begin{tabular}{|l|l|}
			\hline
			Dataset             &                                                Network Architecture                                                 \\ \hline
			DVS Gesture              &   \tabincell{l}{Input-128C5S2-BN-AP2-128C3-BN-AP2-128C3\\-BN-AP2-128C3-BN-AP2-128C3-BN-AP2-512FC\\-VotingC11P5-AP} \\ \hline
			SL-Animals-DVS        &   \tabincell{l}{Input-256C7S2-BN-AP2-256C3-BN-AP2-256C3\\-BN-AP2-256C3-BN-AP2-256C3-BN-AP2-DP\\-512FC-DP-VotingC19P5-AP} \\ \hline
			MNIST-DVS  &                 \tabincell{l}{Input-32C7S2-BN-AP2-64C3-BN-AP2-128C3-BN\\-AP2-512FC-VotingC10P5-AP}              \\ \hline
			
	\end{tabular}}
	
	\label{Tab2}
	
\end{table}
%%%%%%%%%%%%%%%%%%%%%%%%%%

\textbf{DVS Gesture:}~The DVS Gesture dataset contains 11 different gestures recorded from 29 different individuals. These data were captured by an Inilabs DVS camera under three different lighting conditions. In the DVS Gesture dataset, the default numbers of training and test samples are 1176 and 288, respectively.

\textbf{SL-Animals-DVS:}~The SL-Animals-DVS is a more challenging dataset for gesture recognition. Briefly, this dataset includes 1121 samples captured from 58 different subjects performing 19 types of sign language gestures in four collection environments (4 sets). In our experiments, we randomly choose 840 and 281 event streams as training and test samples, respectively. 

\textbf{MNIST-DVS:}~The MNIST-DVS dataset is composed of 30000 DVS handwritten digital data converted from the static MNIST dataset. There are three recording scales~(4, 8 and 16) in the MNIST-DVS dataset, and the total number of event streams for each scale is 10000. Note that we choose the scale of 4 and set the ratio of the training and test sets as 9:1 in our experiments.

%%%%%%%%%%%%%%%%%%%%%%%%%Fig4
\begin{figure*}[t]
	\centering
	\includegraphics[scale=0.95]{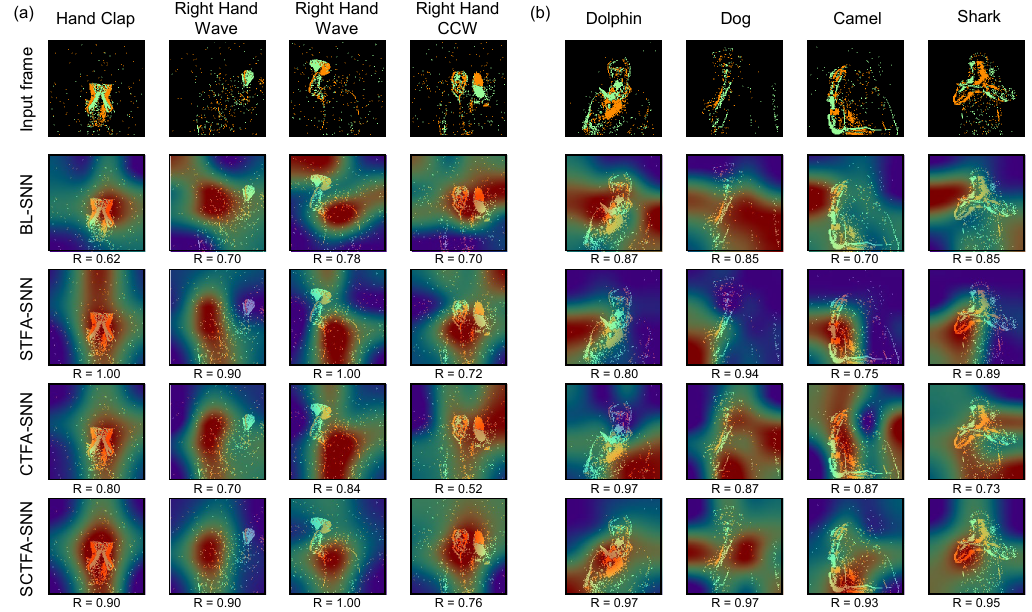}
	\caption{Visualization of the input frames of typical samples and their corresponding attentional heatmaps captured by the BL-SNN, STFA-SNN, CTFA-SNN and  SCTFA-SNN models. (a) DVS Gesture and (b) SL-Animals-DVS. $R$ indicates the average firing ratio in the target class at the voting layer during the inference phase. CCW: Counter Clockwise.}
	\label{fig4}
\end{figure*}
%%%%%%%%%%%%%%%%%%%%%%%%%

%%%%%%%%%%%%%%%%%%%%%%%%%Fig5
\begin{figure*}[h]
	\centering
	\includegraphics[scale=1.01]{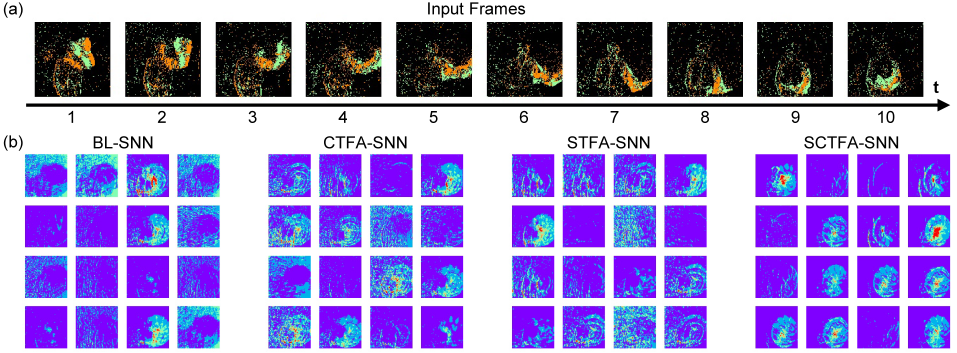}
	\caption{Case study on the DVS Gesture. (a) An example of the input frames of left hand counterclockwise for different timesteps. (b) Visualization of the average spiking activity of neurons for the first 16 channels in the first convolution layer for different SNN models. Red  represents high spiking activation, whereas blue  indicates low spiking activation.}
	\label{fig5}
\end{figure*}
%%%%%%%%%%%%%%%%%%%%%%%%%

%%%%%%%%%%%%%%%%%%%%%%%%%%%%%%%Fig6
\begin{figure*}[t]
	\centering
	\includegraphics[scale=1.02]{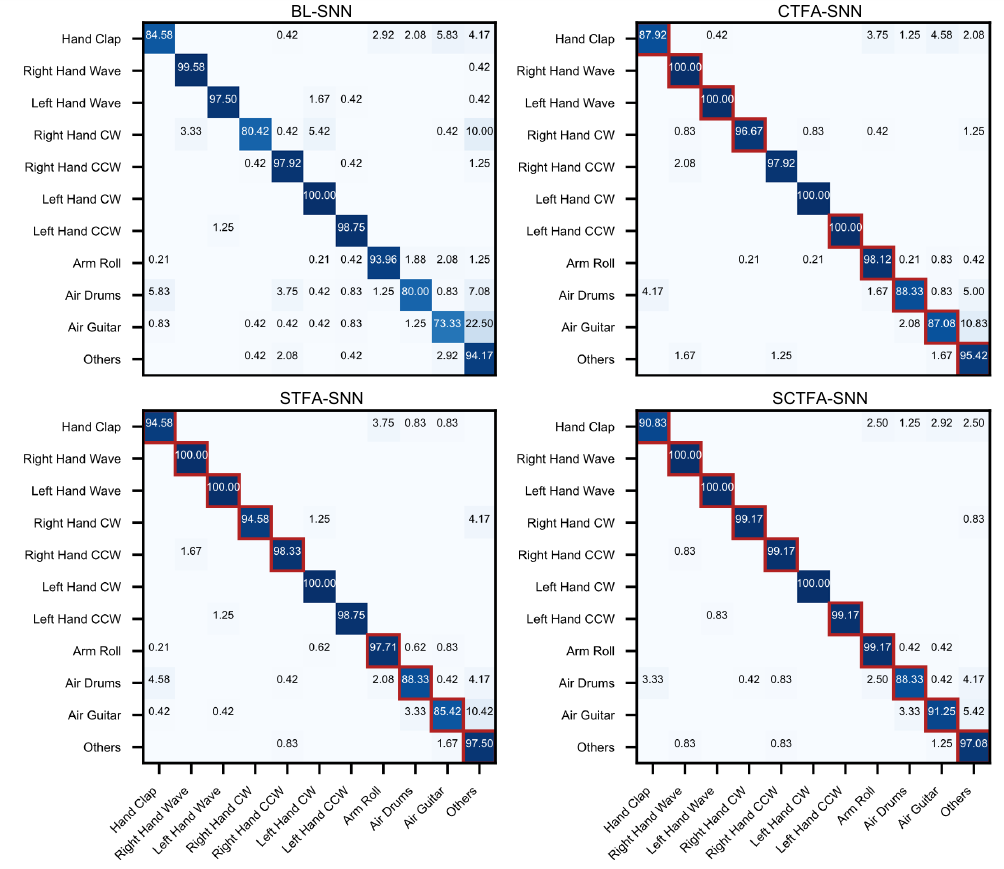}
	\caption{Average confusion matrices of different SNN models on the DVS Gesture.  CW: Clockwise; CCW: Counter Clockwise. The data presented in the confusion matrices are percentages (\%). Compared with the BL-SNN model, the red boxes on the primary diagonal of the confusion matrices denote that the recognition rates of the corresponding gestures are increased.}
	\label{fig6}
\end{figure*}
%%%%%%%%%%%%%%%%%%%%%%%%%%%%%%%	

All experiments are conducted on a workstation equipped with NVIDIA V100 GPUs. The computer codes of our SNN models are implemented by using the PyTorch framework. During the training phase, we combine the Adam optimizer with an exponential decay strategy (initial learning rate~$\eta$ and decay factor $\gamma$). Unless otherwise stated, we use the default hyperparameters listed in Tab.~\ref{Tab1} in our experiments. However, we note that there are a few event streams with their actual lengths less than the required total length ($L=\Delta t \cdot T$). For these event streams, we use the zero-padding method at the end of frame-based inputs to maintain the number of slices as $T$. On different datasets, we construct SNNs with their detailed network architectures given in Tab.~\ref{Tab2}. For example, we train a 7-layer spiking CNN for the SL-Animals-DVS, with its architecture:~[Input-256C7S2-BN-AP2-256C3-BN-AP2-256C3-BN-AP2-256C3-BN-AP2-256C3-BN-AP2-DP-512FC-DP-VotingC19P5-AP]. Here, 256C7S2 refers to the convolutional layer with (channel, kernel~size, stride) = (256, 7, 2), BN represents the batch normalization, and AP2 means a $2\times2$ average pool. DP means dropout with a rate of 0.5. 512FC denotes a fully connected layer with 512 output features. VotingC19P5 indicates that the voting layer has 19 output classes, with 5 neurons for each class. For each experimental setting, we perform 10 independent trials, and report the average (mean $\pm$ standard deviation) and the best top-1 accuracy as the final results.

\subsection{Ablation Study}
We conduct ablation experiments to verify the effectiveness of the proposed method. For a fair evaluation, we train SNNs with different types of attention modules (STFA-SNN, CTFA-SNN, and SCTFA-SNN) and a baseline SNN (BL-SNN) without any attention module on each dataset. In Fig.~\ref{fig3}, we plot the average training curves of different SNN models. For all datasets, the SCTFA-SNN model has relatively fast convergence speed and displays the best average performance. To further compare the accuracy of different SNN models in detail, we summarize the average and best top-1 accuracy over 10 trials in Tab.~\ref{Tab3}. As we see, introducing the historically accumulated channel and spatial information can both improve the baseline performance. However, the STFA-SNN model exhibits slightly higher accuracy than the CTFA-SNN model on two gesture datasets acquired in natural scenes, whereas both models can achieve similar performance on the MNIST-DVS. Further comparisons reveal that the SCTFA-SNN model significantly outperforms these two degenerated SNN models across all datasets. Theoretically, this is  because the SCTFA-SNN model can make full use of the spatial-channel information accumulated in the temporal domain. In particular, we find that the superiority of the SCTFA-SNN is noticeable for tasks that involve complicated spatio-temporal features. In comparison with the BL-SNN model, the average and best top-1 accuracies of the SCTFA-SNN model are improved 6.46\% and 4.86\% on the DVS Gesture, and 5.09\% and 4.63\% on the challenging SL-Animals-DVS [Tab.~\ref{Tab3}]. For the simplest MNIST-DVS, we observe only 1.00\% and 0.70\% improvements in the average and best top-1 accuracies [Tab.~\ref{Tab3}], respectively.

To qualitatively analyze the role of the SCTFA module, we apply Gradient-weighted Class Activation Mapping++ (Grad-CAM++)~\cite{chattopadhay2018grad} to identify the attentional locations for different SNN models. Briefly, Grad-CAM++ is an advanced visualization method that uses a weighted combination of positive gradients to compute the importance of the spatial locations in different convolutional layers~\cite{chattopadhay2018grad}. Figure~\ref{fig4} shows the attentional visualization results of typical samples in the DVS Gesture and SL-Animals-DVS. The red color in the generated heatmap denotes valuable regions captured by the Grad-CAM++ method. Compared with the BL-SNN model, it is obvious that the attention regions of SNNs with different types of attention modules are much more focused on the human body. Furthermore, by computing the average firing ratio $R$ for neurons in the target class at the voting layer, we confirm that the SCTFA-SNN model exhibits considerably larger $R$ values than both the STFA-SNN and CTFA-SNN models, thus indicating its stronger capability to locate targets [Fig.~\ref{fig4}]. 

Moreover, the superiority of the SCTFA module can be seen clearly by observing the average spiking activity of neurons in the convolution layers. In Fig.~\ref{fig5}, we show a case study on DVS Gesture by visualizing the average spiking activity of neurons for the first 16 channels in the first convolution layer. Compared with the other SNN models, we find that the SCTFA-SNN model displays more centralized high-level spiking activation around the underlying target region in several specific channels. This finding implies that the SCTFA module provides a flexible approach to select critical channels by using gradient-based information and greatly reduces redundant spiking activity within these channels, thus resulting in better feature representations.

%%%%%%%%%%%%%%%%%%%%%%%%%%% Table3
\begin{table}[t]
	\renewcommand{\arraystretch}{1.2}
	\caption{Average top-1 accuracy (upper) and best top-1 accuracy (bottom) of the BL-SNN, STFA-SNN, CTFA-SNN and SCTFA-SNN models on different datasets.}
	\centering
	\setlength{\tabcolsep}{1.5mm}{
		\begin{tabular}{|l|c|c|c|}
			\hline
			Model                         &   DVS Gesture    &  SL-Animals-DVS  &    MNIST-DVS     \\ \hline
			\multirow{2}*{BL-SNN}    & $90.87 \pm 1.41\%$         & $81.46 \pm 2.59\%$ & $97.72 \pm 0.25\%$ \\
			~                             &     $93.06\%$                                 &     $85.41\%$      &     $98.20 \%$     \\ \hline		
			\multirow{2}*{STFA-SNN}  & $96.04 \pm 1.28\%$         & $85.44 \pm 1.95\%$ &       $98.30 \pm 0.26\%$      \\
			~                             &     $97.57$\%                               &     $88.97\%$      &      $98.90\%$      \\ \hline
			\multirow{2}*{CTFA-SNN}  & $96.01 \pm 0.64\%$          & $84.56 \pm 1.74\%$ &     $98.40 \pm 0.22\%$      \\
			~                             &     $97.22\%$                                &     $87.90\%$      &      $98.70\%$      \\ \hline
			\multirow{2}*{SCTFA-SNN} & \pmb{$97.33 \pm 0.58\%$}   & \pmb{$86.55 \pm 1.64\%$} & \pmb{$98.72 \pm 0.14\%$} \\
			~                             &     \pmb{$97.92\% $}                         &     \pmb{$90.04\%$}      &     \pmb{$98.90\%$}      \\ \hline
	\end{tabular}}
	\label{Tab3}
	
\end{table}
%%%%%%%%%%%%%%%%%%%%%%%%%%%%%%

For a more insightful analysis, we calculate the average confusion matrices of different SNN models on the DVS Gesture. The corresponding results are illustrated in Fig.~\ref{fig6}. Compared with the results of the BL-SNN model, we find that the accuracies of nine gestures are improved for both the STFA-SNN and CTFA-SNN models. As expected, the SCTFA-SNN model exhibits much higher recognition rates for almost all classes (ten gestures) in comparison with the BL-SNN model, and keeps the 100\% classification accuracy as that of the other SNN models for the ``Left Hand CW'' gesture [Fig.~\ref{fig6}]. Indeed, the SCTFA-SNN model can achieve and maintain the best performance on eight gestures among the compared SNN models, thus leading to a significant improvement in the performance on the DVS Gesture. Note that, in additional experiments, we observe similar results on both the SL-Animals-DVS and MNIST-DVS datasets (data not shown).

Overall, these findings indicate that the SCTFA module can help SNNs to exploit information in appropriate attention regions, thus equipping them with competitive performance for event stream classification tasks. In the following studies, we further analyze both the role of critical hyperparameter and the superiority of the SCTFA-SNN model under different conditions.

\subsection{Impacts of Hyperparameter $\kappa_{\tau}$}

%%%%%%%%%%%%%%%%%%%%%%%%%%%%%%%Fig7
\begin{figure}[t]
	\centering
	\includegraphics{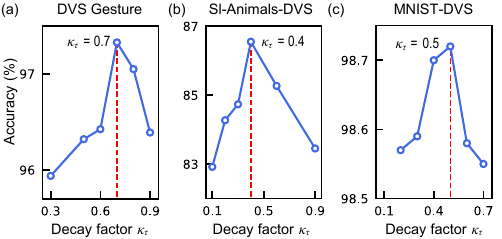}
	\caption{Average top-1 accuracy of the SCTFA-SNN model for different values of hyperparameter $\kappa_{\tau}$ on three datasets. (a) DVS Gesture, (b) SL-Animals-DVS and (c) MNIST-DVS.}
	\label{fig7}
\end{figure}
%%%%%%%%%%%%%%%%%%%%%%%%%%%%%%	

%%%%%%%%%%%%%%%%%%%%%%%%%%%%%%%Fig8
\begin{figure}[t]
	\centering
	\includegraphics{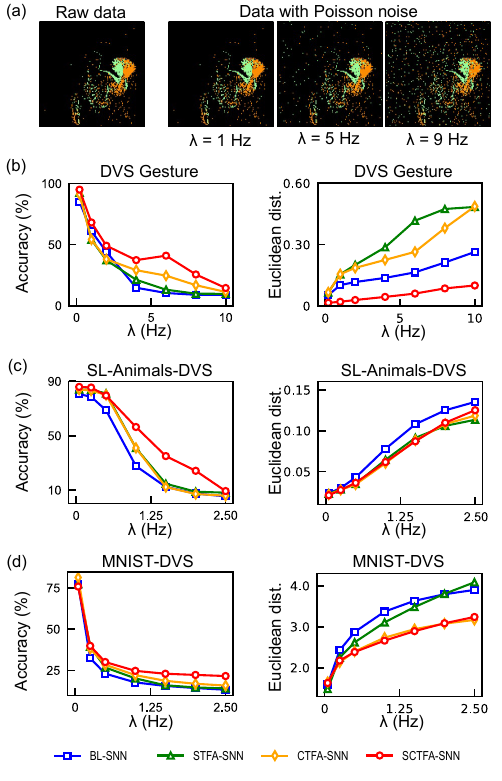}
	\caption{Robustness of different SNN models to Poisson noise on three datasets. (a) An example of frame-based data in the DVS Gesture mixed with different noise levels ($\lambda$). The average accuracy of the BL-SNN (blue), STFA-SNN (green), CTFA-SNN (orange) and SCTFA-SNN (red) models at different noise levels is shown for the DVS Gesture (b), SL-Animals-DVS (c) and MNIST-DVS (d). In addition, the average Euclidean distances between the hidden activation (neuronal membrane potentials in the last convolutional layer) of the original data and those of the noisy data over the timestep are also plotted for different datasets.}
	\label{fig8}
\end{figure}
%%%%%%%%%%%%%%%%%%%%%%%%%%%%%%	

In the SCTFA-SNN model, the decay factor $\kappa_{\tau}$ is a critical hyperparameter that controls the long-term effect of the 3-D attention tensor and impacts the accumulation of spatial-channel information in the temporal direction. A natural question that arises is how the hyperparameter $\kappa_{\tau}$ influences the performance of the SCTFA-SNN model. To address this issue, we carry out comprehensive experiments for the SCTFA-SNN model on three datasets, and the results are presented in Figs.~\ref{fig7}(a)-\ref{fig7}(c), respectively. From a theoretical viewpoint, a very small $\kappa_{\tau}$ means that the historical spatial-channel information captured by the 3D attention tensor cannot be effectively accumulated due to rapid decay. In contrast, an extremely large $k_{\tau}$ dramatically increases the accumulation time, thus introducing a certain amount of early irrelevant or redundant spatial-channel information for locating regions of interest. It is obvious that both of these two cases tend to deteriorate the performance of the SCTFA-SNN model. Consistent with this analysis, our results demonstrate that the SCTFA-SNN model can achieve the best accuracy at an intermediate value of $\kappa_{\tau}$ on all datasets. This finding indicates that accumulating an appropriate amount of historical spatial-channel information, namely, not too much or too little information, can help SNNs by improving their empirical feature representation ability, thus endowing them with superior performance over comparison models. However, we note that the optimal hyperparameter $\kappa_{\tau}$ may differ among various datasets. As shown in Figs.~\ref{fig7}(a)-\ref{fig7}(c), the SCTFA-SNN model can achieve the best performance with $\kappa_{\tau}=0.4$ and 0.5 on the SL-Animals-DVS and MNIST-DVS, whereas a slightly larger value of $\kappa_{\tau}=0.7$ is required to obtain the highest accuracy on the DVS Gesture. We speculate that the change in optimal $\kappa_{\tau}$ value may be caused by multiple factors, such as different target movement speeds and settings of other hyperparameters for these datasets.

\subsection{Robustness to Noisy Data}

%%%%%%%%%%%%%%%%%%%%%%%%%%%%%%%Fig9
\begin{figure}[!t]
	\centering
	\includegraphics{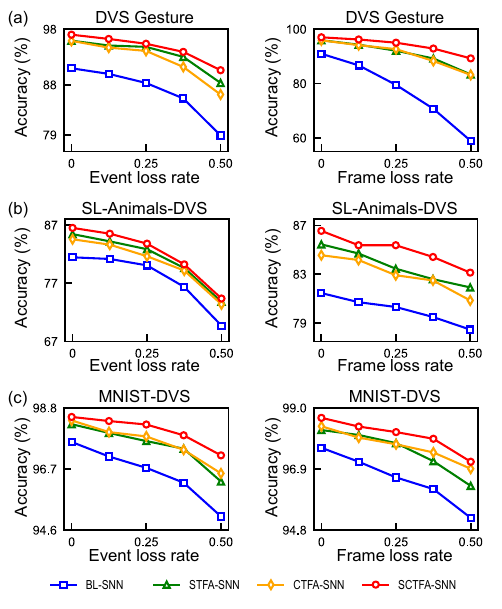}
	\caption{Performance of the BL-SNN (blue), STFA-SNN (green), CTFA-SNN (orange) and SCTFA-SNN (red) models at different event (left) and frame (right) loss rates on three datasets. For both data loss scenarios, we consider that the loss rate is within the range of 0 and 0.50. (a) DVS Gesture, (b) SL-Animals-DVS and (c) MNIST-DVS.}
	\label{fig9}
\end{figure}
%%%%%%%%%%%%%%%%%%%%%%%%%%%%%%

For SNN models, a high robustness to noisy data is essential for ensuring efficient information processing~\cite{wu2022brain}. To evaluate the model robustness to unavoidable noise, we examine the performance of the best pre-trained groups of SNN models (10 independent models) by using noisy data. We assume that the noise satisfies a Poisson distribution, which is believed to be a typical form of noise for event streams~\cite{khodamoradi2018n}. Specifically, we add noisy spiking events governed by a Poisson rate $\lambda$ to event streams in the test set. A larger $\lambda$ represents a higher noise level. In Fig.~\ref{fig8}(a), we schematically show an example of frame-based data in the DVS Gesture mixed with different levels of Poisson noise.

Our results depicted in Figs.~\ref{fig8}(b)-(d) demonstrate that the SCTFA-SNN model shows outstanding robustness to noise in comparison with different SNN models on all three datasets. As the parameter $\lambda$ is increased, we find that the average accuracy of different SNN models basically decreases, and the SCTFA-SNN displays substantial superiority for most cases, especially at intermediate noise levels. However, for extremely high noise levels, the advantage of the SCTFA-SNN model is largely weakened. In reality, this result is not so surprising because useful information represented as sparsely asynchronous events is drowned under strong noise conditions. For a more insightful analysis, we compute the Euclidean distance between the activation of the original data in the last convolutional layer and the activation of noisy data using the same models, and the results are presented in Figs.~\ref{fig8}(b)-(d). Compared with other SNN models, the SCTFA-SNN model tends to reduce Euclidean distance between the noisy patterns and original patterns more significantly for different datasets. These observations suggest that the SCTFA module can help the SNNs fight against stochastic perturbations, thus endowing them with high robustness to noisy data.

\subsection{Stability to Incomplete Data}

Missing data are vital factors that impact the stability of SNN models and can be caused by different reasons. To examine the performance of our model for incomplete data, we consider two typical data loss scenarios: event loss and frame loss. For the case of event loss, we randomly remove sparse events from the original data based on a given event loss rate. For the the case of frame loss, event-based frames are randomly dropped according to a given frame loss rate. Similarly, the best pre-trained group of SNN models is employed to assess their performance on incomplete data. In Figs.~\ref{fig9}(a)-\ref{fig9}(c), we present the average accuracies of different SNN models for both event loss (left) and frame loss (right) on three datasets. As expected, the performance of different SNN models decreases with increasing event and frame loss rates. For all datasets, it can bee seen that the SCTFA-SNN model stably maintains higher accuracy than that of the BL-SNN, STFA-SNN, and CTFA-SNN models for all event and frame loss rates. Compared with the event loss, we identify that the superiority of the SCTFA-SNN model is much more noticeable than the frame loss scenario. This might be because event loss deteriorates the information representation for all input frames, whereas frame loss does not impact the information representations of the data used. Overall, our findings indicate the excellent stability of the SCTFA-SNN model to incomplete data.

%%%%%%%%%%%%%%%%%%%%%%%%%%%%%%Tab4
\begin{table}[t]
	\renewcommand{\arraystretch}{1.2}
	\caption{Complexity and efficiency of the BL-SNN, STFA-SNN, CTFA-SNN and SCTFA-SNN models on different datasets. Here we list the number of parameters, multiply-Adds (Mult-Adds) and actual inference time per batch.}
	\centering
	\setlength{\tabcolsep}{1.0mm}{
		\begin{tabular}{|l|l|c|c|c|}
			\hline
			\multirow{2}*{Dataset}        & \multirow{2}*{Model} &      Params       &     Mult-Adds     & Inference time \\
			~                             & ~                    & ($\times 10^{6}$) & ($\times 10^{9}$) &      (ms)      \\ \hline
			\multirow{3}*{DVS Gesture}    & BL-SNN               &      $0.895$      &      $2.522$      &    $20.100$    \\
			~                             & STFA-SNN             &      $0.896$      &      $2.529$      &    $24.153$    \\
			~                             & CTFA-SNN             &      $0.936$      &      $2.523$      &    $26.422$    \\
			~                             & SCTFA-SNN            &      $0.937$      &      $2.530$      &    $31.778$    \\ \hline
			\multirow{3}*{SL-Animals-DVS} & BL-SNN               &      $2.962$      &     $27.165$      &   $171.262$    \\
			~                             & STFA-SNN             &      $2.963$      &     $27.207$      &   $193.425$    \\
			~                             & CTFA-SNN             &      $3.126$      &     $27.170$      &   $192.663$    \\
			~                             & SCTFA-SNN            &      $3.127$      &     $27.212$      &   $214.424$    \\ \hline
			\multirow{3}*{MNIST-DVS}      & BL-SNN               &      $4.316$      &      $1.096$      &    $71.473$    \\
			~                             & STFA-SNN             &      $4.317$      &      $1.101$      &    $81.642$    \\
			~                             & CTFA-SNN             &      $4.327$      &      $1.096$      &    $80.142$    \\
			~                             & SCTFA-SNN            &      $4.327$      &      $1.101$      &    $90.481$    \\ \hline
	\end{tabular}}
	\label{Tab4}
\end{table}
%%%%%%%%%%%%%%%%%%%%%%%%%%%

%%%%%%%%%%%%%%%%%%%%%%%%%%%%%%Tab5
\begin{table*}[!t]
	\renewcommand{\arraystretch}{1.2}
	\newcommand{\tabincell}[2]{
		\begin{tabular}
			{@{}#1@{}}#2
		\end{tabular}
	}
	\caption{A detailed comparison among the SCTFA-SNN model and other existing models on different datasets.} %IF: Integrate-and-fire;  LIF: Leaky Integrate-and-fire; ReLU: Rectified Linear Unit; GELU: Gaussian Error Linear Unit; SRM: Spiking Response Model}
	\centering
	\begin{threeparttable}  
		\setlength{\tabcolsep}{2.3mm}{
			\begin{tabular}{|c|l|c|c|c|c|c|}
				\hline
				Dataset                            & Model                                                            &       Neuron type        &         Network architecture         &        Method         &    $\Delta t \times T$     &              Accuracy              \\ \hline
				\multirow{19}*{\rotatebox{0}{\rotatebox{90}{DVS Gesture}}}  & Zheng et al. (2021)~\cite{zheng2021going}                        &           LIF            &               ResNet17               &         STBP          &    $30\times40$     &             $96.87\%$              \\
				& Fang, et al. (2021)~\cite{fang2021incorporating}                 &           PLIF           &         7-Layer Spiking CNN    &         STBP          & $\times20$\tnote{1} &             $97.57\%$              \\
				& Wu et al. (2021) ~\cite{wu2021liaf}                              &           LIAF           &         5-Layer Spiking CNN          &         BPTT          &    $25\times60$     &             $97.56\%$              \\
				& Yao et al. (2021)~\cite{yao2021temporal}                         &           LIF            &         5-Layer Spiking CNN          &       STBP + TA       &    $15\times60$     &             $95.49\%$              \\
				& Yao et al. (2021)~\cite{yao2021temporal}                         &           LIAF           &         5-Layer Spiking CNN          &       BPTT + TA       &    $25\times60$     &             $98.61\%$              \\
				& Sun et al. (2022)~\cite{sun2022synapse}                          &           LIF            &         7-Layer Spiking CNN          &       STL-STBP        &     $\times20$\tnote{1}      &             $97.22\%$              \\
				& Baldwin et al. (2021) ~\cite{baldwin2022time}                    &           GELU           &              GoogLeNet               &          BP           &     $25\times9$     &             $96.20\%$              \\
				& Fang et al. (2021) ~\cite{NEURIPS2021afe43465}  &  LIF   & 7B-Net~(SEW ResNet) & STBP &     $\times16$\tnote{1}      &     $97.92\%$     \\
				& Feng et al. (2022) ~\cite{ijcai2022p343}        & MLF-LIF &  Spiking DS-ResNet  & STBP &    $30\times40$     &     $97.29\%$     \\
				& Zhu et al. (2022) ~\cite{zhu2022tcja}           &   LIF   & 7-layer Spiking CNN & STBP &     $\times20$\tnote{1}      &  \pmb{$99.0\%$}   \\
				& Yu et al. (2022) ~\cite{103389fnins20221079357} &   LIF   & 7-layer Spiking CNN & STBP &     $\times20$\tnote{1}      &  \pmb{$98.96\%$}  \\
				& Sabater et al. (2022) ~\cite{sabater2022event}                   &           ReLU           &                 EvT                  &          BP           &          --          &             $96.20\%$              \\
				& Shrestha et al. (2018) ~\cite{shrestha2018slayer}                &           SRM            &         8-Layer Spiking CNN          &        SLAYER         &    $5\times300$     &         $93.64 \pm 0.49\%$         \\
				& Kugele et al. (2020) ~\cite{kugele2020efficient}                 &            IF            &               DenseNet               &        ANN-SNN        &   $6.25\times240$   &         $95.56 \pm 0.14\%$         \\
				& Zhu et al. (2021)  ~\cite{zhu2021efficient}                      &          C-LIF           &                LeNet                 &         STBP          &    $1\times500$     &         $95.83 \pm 0.28\%$         \\
				& Sun et al. (2022)  ~\cite{sun2022synapse}                        &           LIF            &         7-Layer Spiking CNN          &       STL-STBP        &     $\times20$\tnote{1}      &         $96.84 \pm 0.24\%$         \\
				& Wu et al. (2022)  ~\cite{wu2022brain}                            &           LIF            &         9-Layer Spiking CNN          &          HP           &    $10\times40$     &         $97.01 \pm 0.21\%$         \\
				& \pmb{Our work}\tnote{2} (Best top-1)                            &           LIF            &         7-Layer Spiking CNN          &     STBP + SCTFA      &    $125\times10$    &             $97.92\% $             \\
				& \pmb{Our work}\tnote{2} (Average top-1)                          &           LIF            &         7-Layer Spiking CNN          &     STBP + SCTFA      &    $125\times10$    &         $97.33 \pm 0.58\%$         \\
				& \pmb{Our work}\tnote{3}  (Best top-1)                           &           LIF            &         7-Layer Spiking CNN          &     STBP + SCTFA      &    $125\times20$    &         \pmb{$98.96\% $ }          \\
				& \pmb{Our work}\tnote{3}~(Average top-1)                          &           LIF            &         7-Layer Spiking CNN          &     STBP + SCTFA      &    $125\times20$    &      \pmb{$98.44 \pm 0.39\%$}      \\ \hline
				\multirow{7}*{\rotatebox{0}{\rotatebox{90}{SL-Animals-DVS}}} & Baldwin et al. (2021)  ~\cite{baldwin2022time}                   &           GELU           &              GoogLeNet               &          BP           &     $25\times9$     &             $85.10\%$              \\
				& Sabater et al. (2022)   ~\cite{sabater2022event}                 &           ReLU           &                 EvT                  &          BP           &          --          &             $88.12\%$              \\
				& Vasudevan et al. (2020) ~\cite{vasudevan2020introduction}        &           SRM            &         4-Layer Spiking CNN          &         STBP          &    $30\times50$     &         $56.20 \pm 1.52\%$         \\
				& Vasudevan et al. (2020) ~\cite{vasudevan2020introduction}        &           SRM            &         4-Layer Spiking CNN          &        SLAYER         &    $5\times300$     &         $60.09 \pm 4.58\%$         \\
				& Vasudevan et al. (2022) ~\cite{vasudevan2022sl}                  &           SRM            &         5-Layer Spiking CNN          &        DECOLLE        &    $1\times500$     &         $70.60 \pm 7.80\%$         \\
				& \pmb{Our work}\tnote{2} (Best top-1)                                      &           LIF            &         7-Layer Spiking CNN          &     STBP + SCTFA      &    $50\times30$     &          \pmb{$90.04 \%$}          \\
				& \pmb{Our work}\tnote{2} (Average top-1)                                   &           LIF            &         7-Layer Spiking CNN          &     STBP + SCTFA      &    $50\times30$     &      \pmb{$86.55 \pm 1.66\%$}      \\ \hline
				\multirow{9}*{\rotatebox{0}{\rotatebox{90}{MNIST-DVS}}}    & Paulun et al. (2018)  ~\cite{paulun2018retinotopic}              &           LIF            &         Retinotopic Mapping          &       STDP + BP       &    $10\times100$    &        $90.56\%$~(scale-4)         \\
				& Cannici et al. (2019) ~\cite{cannici2019attention}               &           LSTM           &           4-Layer ConvLSTM           &         BPTT          &          --          &        $98.30\%$~(scale-4)         \\
				& Sun et al. (2022)    ~\cite{sun2022synapse}                      &           LIF            &         7-Layer Spiking CNN          &       STL-STBP        &     $\times20$\tnote{1}      &        $98.70\%$~(scale-4)         \\
				& Wang et al. (2021)  ~\cite{wang2021compsnn}                      &            IF            &         4-Layer Spiking CNN          &       Tempotron       &          --          &        $82.65\%$~(scale-16)        \\
				& Liu et al. (2022)   ~\cite{liu2022event}                         &           LIF            &         4-Layer Spiking CNN          &         STBP          &     $\times64$\tnote{1}      &             $96.64\%$~             \\
				& Zhu et al. (2021)  ~\cite{zhu2021efficient}                      &          C-LIF           &                LeNet                 &         STBP          &    $100\times25$    &    $98.40 \pm 0.27\%$~(scale-4)    \\
				& Sun et al. (2022)  ~\cite{sun2022synapse}                        &           LIF            &         7-Layer Spiking CNN          &       STL-STBP        &     $\times20$\tnote{1}      &    $98.37 \pm 0.17\%$~(scale-4)    \\
				& \pmb{Our work}\tnote{2} (Best top-1)                                      &           LIF            &         5-Layer Spiking CNN          &     STBP + SCTFA      &    $25\times20$     &     \pmb{$98.90\%$}~(scale-4)      \\
				& \pmb{Our work}\tnote{2} (Average top-1)                                   &           LIF            &         5-Layer Spiking CNN          &     STBP + SCTFA      &    $25\times20$     & \pmb{$98.72 \pm 0.14\%$}~(scale-4) \\ \hline
				
		\end{tabular}}
		\begin{tablenotes} 
			\footnotesize
			\item[1] Each event stream is uniformly divided into $T$ slices based on the number of events. 
			\item[2] Performance of the SCTFA-SNN model with the default experimental setup given in Tab.~\ref{Tab1}.
			\item[3] Performance of the SCTFA-SNN model with the same experimental setup as used for the TCJA-SNN and STSC-SNN models~\cite{zhu2022tcja, 103389fnins20221079357}, while keeping the slice length $\Delta t=125$~ms.
		\end{tablenotes}
	\end{threeparttable}
	\label{Tab5}
\end{table*}
%
%%%%%%%%%%%%%%%%%%%%%%%%%%%

\subsection{Analysis of Complexity and Efficiency}
Introducing the SCTFA module into SNNs results in a trade-off between the cost and performance. To further analyze the model complexity and efficiency, we calculate the numbers of parameters, multiply-add operations (Mult-Adds) and actual inference time per batch for the BL-SNN and SNNs with different attention modules. As comparisons shown in Tab.~\ref{Tab4}, the SCTFA module introduces slightly more parameters and computational complexity into SNNs than the CTFA and STFA modules. For SNNs with different attention modules, such enhancement of computational complexity leads to a relatively long, but reasonable, increase in the actual inference time. This non-negligible efficiency loss may be attributed to the increased the number of element-wise operations and reduced parallelism degree due to the insertion of attention modules. During the inference process, we find that the STFA-SNN and CTFA-SNN models exhibit  similar inference times on all datasets. In comparison with the BL-SNN model, the most complicated SCTFA-SNN model spends an extra inference time of 11.678~ms, 43.162~ms and 19.008~ms per batch on the DVS Gesture, SL-Animals-DVS and MNIST-DVS, respectively [Tab.~\ref{Tab4}]. Considering the length of event stream data used in this study, we believe that the realistic runtime of the SCTFA-SNN model is still acceptable for real applications. Overall, our above analysis suggests that the proposed SCTFA-SNN model can achieve superior performance, while maintaining a certain degree of competitiveness in terms of complexity and efficiency.

\subsection{Comparison with Prior SOTA Methods}
Finally, we compare the proposed SCTFA-SNN model against other existing SOTA works. For a fair comparison, we mainly include results of SNNs with signal transmission via binary spikes, but exclude those with data augmentation. A comparison of accuracy among various works is summarized in Tab.~\ref{Tab5}. Obviously, the SCTFA-SNN model can achieve the SOTA results on both the SL-Animals-DVS (average: $86.55 \pm 1.66\%$; best: 90.04\%) and MNIST-DVS (average: $98.72 \pm 0.14\%$; best: 98.90\%). In particular, we get a considerable increase of 15.95\% on average over the prior best SNN model on the SL-Animals-DVS~\cite{vasudevan2022sl}, and our SCTFA-SNN model even outperforms the accuracy of the current best non-SNN method, known as event transformer (EvT), by 1.92\%~\cite{sabater2022event}. On the DVS Gesture, our best result (97.92\%) obtained by default experimental setup is slightly lower than the results of several models reported in recent studies. For instance, it has been shown that the TA-SNN model can achieve a higher accuracy of 98.61\% on the DVS Gesture~\cite{yao2021temporal}. However, this SOTA accuracy is achieved by not only using a relatively larger timestep of $T=60$ but also utilizing the leaky integrate and analog fire (LIAF) neurons~\cite{wu2021liaf, yao2021temporal}, which transmit signal via analog values but not binary spikes. Indeed, the performance of our model is better than that of the corresponding SNN with spiking LIF neurons (95.49\%), which is also reported in the same work [see Tab.~\ref{Tab5}]~\cite{sabater2022event}. On the other hand, it seems that our default SCTFA-SNN model has a small performance gap compared with the TCJA-SNN~\cite{zhu2022tcja} and STSC-SNN~\cite{103389fnins20221079357} models. However, we postulate that this gap might be mainly because both TCJA-SNN and STSC-SNN use different experimental setups during the learning and inference phases, including a cosine annealing method to control the learning rate, the detachment of the reset process during backpropagation, a much larger number of training epochs (1000), and a larger timestep of $T=20$~\cite{zhu2022tcja, 103389fnins20221079357}. By introducing the same experimental setup as above while keeping the default slice length~($\Delta t=125$~ms, see Tab.~\ref{Tab1}), our experimental results demonstrate that the average and best top-1 accuracies of the SCTFA-SNN model can be further increased to 98.96\% and $98.44 \pm 0.39\%$, thus achieving the same performance (best top-1) as the TCJA-SNN and STSC-SNN models. In reality, this finding is quite surprising because only a short  event stream  ($\Delta t \cdot T = 2500$~ms) is used in the SCTFA-SNN model, whereas the TCJA-SNN and STSC-SNN models are trained and tested with complete event streams (average data length: 6531.5~ms). These results imply that our SCTFA-SNN model is competitive with other existing methods, and the proposed SCTFA module can remarkably improve the performance of SNNs by combining the spatial, channel and temporal information.

\section{Conclusions}
As an emerging field in artificial intelligence, SNNs have attracted increasing attention and promoted neuromorphic computing in recent years~\cite{maass1997networks, ghosh2009spiking, tavanaei2019deep, kim2020spiking, rast2018behavioral, bing2018survey}. There is a broad consensus that the development of highly efficient SNN models should adequately learn from brain mechanisms~\cite{cheng2020lisnn, pei2019towards, hong2019training}. Visual attention is essential for human perception~\cite{carrasco2011visual, lamme2003visual}. Recently, a few studies devoted to introduce attention mechanisms into SNNs to extract important information in spatial, temporal and channel dimensions separately~\cite{yao2021temporal, kundu2021spike, liu2022event, liu2022general}, but almost none of them have considered to comprehensively combine the information from all dimensions together. In this work, we designed a friendly plug-and-play SCTFA module for SNNs based on the idea of the predictive attentional remapping~\cite{melcher2007predictive, rolfs2011predictive, mathot2010evidence, szinte2018pre, wilmott2021transsaccadic, golomb2021visual}. In contrast to other attention mechanisms that have been inserted in previous SNN models, the main novelty of our approach is that the proposed SCTFA module allows SNNs to make full use of historical spatial-channel information accumulated in the temporal domain to efficiently predict current attention regions. Through experiments, we identified that the SCTFA-SNN model not only achieves higher accuracy than two other SNN models with degenerated spatial-temporal-fused and channel-temporal-fused attention modules, but also shows the competitive performance with existing SOTA models on various event stream datasets. In addition, our analysis showed that our SCTFA-SNN model exhibits excellent robustness and stability under different conditions, despite increasing a certain degree of complexity and inference time.

Our findings presented in this study emphasize that  incorporating appropriate brain-inspired computational principles can improve the information processing capability of SNNs. In future studies, we plan to train the SCTFA-SNN model with more efficient hybrid or synergistic learning strategies~\cite{wu2022brain, sun2022synapse}. Moreover, we hope to further generalize the SCTFA-SNN model to other challenging intelligent applications and  
implement this model on neuromorphic chips~\cite{pei2019towards, akopyan2015truenorth, schuman2022opportunities, ma2017darwin,  8668700,  9142407}. Finally, we note that some of our choices in the present study, such as the SNN architecture, which was adapted based on the traditional ANN, frame-based inputs and rate-based decoding strategy, result in an SNN model that is distinct from its biological basis. In future studies, these shortcomings should be addressed by introducing more biological constraints and mechanisms.

\section*{Acknowledgments} We sincerely thank Dr. Xuhui Huang for valuable comments and discussions on the preliminary results of this work. We are grateful to Prof. Liangjian Deng, Ruijie Zhu and Chengting Yu for providing implementation details of the TCJA-SNN and STSC-SNN models.

\bibliographystyle{IEEEtran}
\bibliography{reference.bib}

\end{document}